# Automatic Identification of Stone-Handling Behaviour in Japanese Macaques Using LabGym Artificial Intelligence

Théo Ardoin[1], Cédric Sueur[2,3,4]

1 Magistère de Biologie, Université Paris-Saclay, Paris, France

2 Université de Strasbourg, IPHC UMR7178, CNRS, Strasbourg, France5

3 ANTHROPO-LAB, ETHICS EA 7446, Université Catholique de Lille, Lille, France

4 Institut Universitaire de France, Paris, France

**Abstract:** The latest advancements in artificial intelligence technology have opened doors to the analysis of intricate behaviours. In light of this, ethologists are actively exploring the potential of these innovations to streamline the time-intensive process of behavioural analysis using video data. In the realm of primatology, several tools have been developed for this purpose. Nonetheless, each of these tools grapples with technical constraints that we aim to surmount. To address these limitations, we have established a comprehensive protocol designed to harness the capabilities of a cutting-edge tool, LabGym. Our primary objective was to evaluate LabGym's suitability for the analysis of primate behaviour, with a focus on Japanese macaques as our model subjects. We have successfully developed a model that demonstrates a high degree of accuracy in detecting Japanese macaques stone-handling behaviour. Our behavioural analysis model was completed as per our initial expectations and LabGym succeed to recognise stone-handling behaviour on videos. However, it is important to note that our study's ability to draw definitive conclusions regarding the quality of the behavioural analysis is hampered by the absence of quantitative data within the specified timeframe. Nevertheless, our model represents the pioneering endeavour, as far as our knowledge extends, in leveraging LabGym for the analysis of primate behaviours. It lays the groundwork for potential future research in this promising field.

**Introduction**

The study of animal behaviour is essential for understanding the ecological niche of a species, evolution of its behaviour, etc. (Barnard 2012). To conduct an ethological study, it is necessary to precisely identify the individuals under investigation and characterise their actions. Traditionally, direct observation of wild animals in their natural habitat has been employed to collect behavioural data (Altmann 1974; Huntingford 2012). However, this approach has limitations due to the real-time precision of observations and the disruptive effect of human presence on the behaviour of the studied species. Indeed, the degree of habituation to human presence varies considerably among species and individuals, thereby introducing numerous observation biases.

With the advent of handheld cameras, ethological field studies now focus on capturing videos that are subsequently analysed in the laboratory. This approach offers several advantages, including the possibility of delayed analysis, observation without human disturbance, and the creation of citizen science databases, which are valuable for research and conservation. Nevertheless, studying complex behaviours from video recordings remains challenging for untrained human observers. Moreover, the analysis of extensive video data is extremely time-consuming and demands substantial human resources. For instance, it is estimated that a detailed analysis of a video takes, on average, three times its duration (Anderson and Perona 2014).

The evolution of computer technologies has ushered in new possibilities for image and video processing through the use of artificial intelligence (Kabra et al. 2013; Valletta et al. 2017; Chakravarty et al. 2020; Kleanthous et al. 2022). The early 2010s marked a significant turning point in computing with the widespread availability of graphics processing units (GPUs), which are computer components capable of simultaneously handling multiple heavy calculations. This newfound capability enables our computers to support a sophisticated technology known as Convolutional Neural Networks (CNNs) (Gu et al. 2018; De Cesarei et al. 2021 ; Shi et al. 2022). CNNs are artificial neural networks whose architecture is inspired by the organisation of visual neurons in mammals. The depth of a CNN, determined by the number of layers of neurons it comprises, is proportional to its capacity for complex analysis – referred to as 'network depth'. This brain-inspired structure allows CNNs to learn from training data (Roffo 2017). This technological advancement paved the way for the development of the first tracking or automatic monitoring software, such as Ethovision® (Noldus et al. 2001), marking the inception of computer-based animal behaviour analysis, also known as computational ethology. This software offers precise and objective means of studying animals by quantitatively analysing various spatiotemporal parameters that define their kinematics. As a result, it represents a substantial time-saving tool for researchers, enabling them to simultaneously collect and analyse behavioural data through computer input. Unfortunately, at that time, common computers were not equipped to support CNNs with significant complexity. Consequently, tracking software faced limitations in terms of learning performance and detection quality, making them less versatile tools. Primarily, they were designed for studying laboratory animals against a neutral background, meeting the highest demand in controlled settings. However, due to limited learning capacity, recognising new animals or behaviours often required manual source code updates by developers. Furthermore, the analysis of an animal's kinematics was not comprehensive enough to identify all behaviours. Imperfect animal detection often necessitated filming animals from above. While this angle simplified AI analysis, it inevitably resulted in a loss of information, such as the inability to quantify leg movements, for example. These limitations made computational ethology (Anderson and Perona 2014) impractical for studying wildlife.

A technical revolution in 2015 changed the landscape with the introduction of the first Residual Neural Network (ResNet) (Liang 2020). This innovation addressed a common challenge faced by the scientific community working on deep CNNs – the vanishing gradient problem. This problem refers to the AI's difficulty in correcting its predictions, which increases exponentially as the number of layers in the artificial neural network grows. The ResNet architecture successfully overcame this issue, preventing a significant decrease in CNNs' learning capacity as their complexity increased. As a result, the first computational ethology algorithms inspired by ResNet emerged, providing versatile tools with continuously improving learning capabilities. The majority of AI developed during this period was designed for supervised training under the guidance of human experts. Consequently, from 2015 to 2020, numerous online annotation tools were developed to assist humans in image and video analysis. In the field of ethology, particularly in primatology, a notable example is the 'MacaquePose' project (Labuguen et al. 2021), which was completed in January 2021. In this project, a team of researchers used a database containing over 13,000 annotated images of macaques to train DeepLabCut AI for postural analysis (Hardin and Schlupp 2022; Lauer et al. 2022). Achieving the accuracy of over 90%, this AI can determine the posture of macaques in each image extracted from a video. This high precision enables the monitoring of macaques in their natural environment, representing a crucial first step in behavioural analysis.

However, one significant challenge persists, as noted by the authors of the article: the limitation of a 2-dimensional view when analysing macaque behaviour. The reduction to two dimensions may result in a potential loss of information regarding behaviours. To address this issue, two potential solutions are being considered. First, there is the option of using a 3-dimensional video capture approach employing multiple cameras. This method provides the most accurate recording of primate behaviour but demands substantial material resources, making it less applicable for studying primates in their natural habitat. The second approach involves enhancing two-dimensional image processing models, which is favoured for practical reasons and holds great promise for the scientific community. Many research teams are opting to integrate a new type of algorithm into their artificial intelligence architecture known as Mask R-CNN (Bharati and Pramanik 2020; Xu et al. 2020, 2021). Currently, this represents one of the most advanced computer vision algorithms available. Notably, Mask R-CNN enables 'segmentation by instance' of the objects being analysed. This term refers to the AI's ability to precisely distinguish each instance of the same object in an image, a task that was previously challenging for classic CNNs. As a result, AI systems can now detect and differentiate several individuals as distinct entities, marking a significant advancement in computational ethology applied to wildlife research (Schofield et al. 2019; Charpentier et al. 2020 ; Rigoudy et al. 2022; Paulet et al. 2023).

LabGym1.0, as introduced in reference (Hu et al. 2023), represents a significant milestone in the field of computational ethology. This software empowers the general public to train artificial neural networks for animal identification and behaviour analysis based on laboratory or natural environment video footage. What sets LabGym1.0 apart from its competitors is its exceptional accessibility. First, LabGym1.0 distinguishes itself by not requiring a GPU for operation, making it cost-effective. This accessibility aligns with an inclusive approach to computational ethology, particularly relevant in fields like primatology, where many nature reserves are situated in economically developing countries. In such settings, data analysis often takes place on-site, where limited resources can be transported. By eliminating the GPU requirement, LabGym1.0 addresses this challenge. Furthermore, LabGym1.0 stands out as it was designed with users possessing limited computer skills in mind. Its user-friendly interface eliminates the need for coding expertise, except during the initial installation process.

This software choice was deliberate in our project, which aims to create a model for studying the behaviour of Japanese macaques (Nakagawa et al. 2010) using artificial intelligence. Our project serves a dual purpose. Firstly, it forms part of fundamental research, assessing the feasibility of developing such a model given the complexity of behavioural identification. Secondly, it aligns with a long-term objective of creating freely accessible AI tools capable of analysing the individual and social behaviours of various primate species. This work has the potential to complement automatic identification models for primates, forming a comprehensive module for in-depth analysis of primate social and cultural behaviours. Indeed, studying and understanding social and cultural behaviours is important for the conservation of primates (Naud et al. 2016; Brakes et al. 2019; Romano et al. 2020; Sueur et al. 2023).

The creation of our model proceeded through three distinct stages. Initially, we trained an image analysis AI to identify Japanese macaques within the video frames. This step enabled us to acquire the precise position coordinates of each individual present in the video images. Utilising these position coordinates, we then proceeded to train an artificial neural network capable of recognising and classifying the behaviours exhibited by these detected primates. Finally, we conducted an assessment of the AI's predictions to gauge its accuracy. Throughout this project, our focus was on the study of stone-handling behaviours, a valuable avenue for investigating culture within the Japanese macaque population. Our choice was motivated by the sheer diversity of stone game behaviours observed, encompassing a total of 45 distinct behaviours. Additionally, the extensive existing literature on these behaviours provided a solid foundation for our research (Leca et al. 2007, 2012; Huffman et al. 2010; Nahallage et al. 2016).

Our model serves as a foundational framework that can be expanded upon in future work, exploring other behaviours of interest. In the long term, we envision integrating our model with an automatic identification system for individual macaques within a group (Paulet et al. 2023). This holistic approach will enable comprehensive studies of both individual and sociocultural behaviours among Japanese macaques and potentially other primate species, as well as non-primate species, broadening the scope of ethological research.

**Material & Methods**

*Study subjects: Japanese macaques (Macaca fuscata)*

The Japanese macaque holds the distinction of being the northernmost primate species (Nakagawa et al. 2010). Endemic to Japan, it thrives in diverse ecosystems and forms matrilineal societies. This species was among the earliest subjects of study in the field of primatology, dating back to 1950 (Matsuzawa and McGrew 2008). Since then, Japanese Macaques have enjoyed protected status in Japan, where they are the focus of research both in reserves and in their natural habitats. The variety of environments they inhabit provides an opportunity to examine their behaviour in relation to environmental factors. With 73 years of dedicated research, the scientific community has accumulated a wealth of data and knowledge about these primates (Nakagawa et al. 2010). Notably, it was among Japanese macaques that cultural transmission was first observed in a non-human animal. This term describes a scenario where young individuals learn behaviours by observing their elders, often their parents (Whiten 2021). This phenomenon has been the subject of longitudinal studies spanning more than three decades, leading to the identification of numerous cultural behaviours within the Japanese Macaque population. One of these behaviours, the stone-handling behaviour has garnered increasing interest.

*Study Behaviour: Stone-Handling or Stone Playing*

Stone's handling or stone play is a solitary activity observed in various species of macaques, including the Japanese macaque (Leca et al. 2007, 2012; Huffman et al. 2010; Nahallage et al. 2016). This term encompasses a collection of 45 different behaviours, each observed with varying frequencies within different groups. The frequencies of these stone game behaviours within a group collectively define the 'stone game culture' of that particular group. Notably, the culture of stone play is transmissible through observation.

For instance, specialists in Japanese macaque behaviour have demonstrated that when a foreign male joins a group, it often brings stone-playing behaviours from its original group into the new one, illustrating a cultural influence. Additionally, environmental factors also influence stone play in Japanese Macaques. A comprehensive comparative study examined various populations of Japanese Macaques across Japan, enabling researchers to assess the impact of environmental factors on the practice of stone play (Leca et al. 2007, 2008, 2011). Among the climatic factors studied, researchers found a positive correlation between stone play and factors like sunshine and high temperatures. Furthermore, stone play appears primarily in populations of macaques that are fed by humans. This is because when macaques are provided with food, they spend less time foraging, which leaves them with more leisure time. This surplus free time is then often used for exploring their environment and engaging in enjoyable stone games.

In some groups, stone play has been observed for over three decades. Beyond its recreational aspect, it appears to play a significant role in the development and maintenance of the sensorimotor cortex of macaques (Nahallage et al. 2016). As a result, this behaviour offers a clear evolutionary advantage. Stone's play has become an integral part of the daily lives of these macaques, and it aligns with a popular hypothesis in primatology, suggesting that the regular handling of stones may eventually lead to their use as tools. Therefore, stone play serves as both a cultural phenomenon and a potential stepping stone toward the development of new tools. Studying this behaviour sheds light on the adaptations of Japanese macaques to their environment, offering insights into their individual and species-level adjustments.

*Analysis Software: Labgym*

LabGym1.0 is an open-source software designed for both qualitative and quantitative analysis of animal behaviours. With LabGym1.0, users have the flexibility to create artificial neural networks (ANNs) whose complexity they can customise (Enquist and Ghirlanda 2005). The complexity of an Artificial Neural Network, or ANN, is primarily determined by its depth (the number of layers it comprises) and its width (the size of each layer). This customisation allows users to adjust parameters according to their specific needs, enabling the creation of shallow ANNs suitable for tasks like image analysis or simple behaviour analysis, which in turn allows for quicker analysis. For an AI to accurately analyse an individual's movements, it must possess the capability to detect the individual with a high degree of confidence. In the context of LabGym1.0, it is essential to isolate the animal from the background to focus the analysis accurately. To tackle this challenge, LabGym1.0 provides users with the ability to train two types of artificial intelligence.

The first type is a 'Detector,' an AI whose architecture employs Detectron2 technology (Jabir et al. 2021), which is image recognition software developed by Meta, the parent company of Facebook. Once trained, this AI becomes proficient at identifying the set of pixels within an image that represents an animal. Consequently, the detector can determine the animal's location within the image as well as its posture.

It's worth noting that the LabGym1.0 version available online until 07/12/2023 (the release date of LabGym2.0) is limited to detecting a single class of objects simultaneously. For instance, it can simultaneously detect ten Japanese macaques in an image because they belong to the same class (Japanese macaque). However, it cannot detect both a stone and a macaque during the same analysis, as they are not classified under the same object class. Therefore, this project must focus on the analysis of a single class of objects.

The position and posture of individuals within each image that comprises a video are then processed by another artificial intelligence known as a 'Categoriser.' This AI utilises the information provided by the detector and analyses it to identify specific actions or behaviours. The Categoriser employs a ResNET-type architecture (Liang 2020), capable of handling complex analyses necessary for extracting spatiotemporal phenomena from a series of images. Training the Categoriser, similar to the detector, requires robust and rigorous training.

*Artificial Intelligence Training*

The two artificial intelligences employed by LabGym1.0 require supervised training under the guidance of a human. The training process begins with the detector, followed by training the Categoriser. In supervised training, human-provided examples that are validated and verified are used to instruct the machine (Gris et al. 2017 ; Valletta et al. 2017; Bohnslav et al. 2021).

*Training the Detector:* In this initial phase, the Detector AI is trained to recognise and locate the target objects or animals within images. Human supervisors provide a set of annotated examples where they identify and label the positions of the objects of interest in the images. The AI learns from these annotated examples and uses this knowledge to detect similar objects in new, unlabelled images. This process continues until the detector can accurately and reliably detect the specified objects. The primary goal of our detector is to identify Japanese macaques within an image, which involves precisely selecting all the pixels representing an individual macaque. This typically involves outlining a perimeter on the image that encompasses all relevant pixels, a process referred to as 'annotation'. To train the detector effectively for Japanese macaque recognition, we need to annotate numerous macaques in various environmental settings and positions to ensure its versatility. Among the current methods for training artificial intelligence to analyse images, annotation is widely used. Several free online tools have been developed to simplify and streamline the annotation process. For our project, we have chosen to utilise the Roboflow® platform (Lin et al. 2022; Paulet et al. 2023) due to its compatibility with datasets used by other research teams working on Japanese macaques. It's worth noting that there are other equally capable tools, such as VGG Annotator® (Dutta et al. 2016), that can successfully perform this annotation task as well. By annotating macaques in a diverse range of situations, we can ensure that our detector is well equipped to recognise them accurately in different environments and positions, making it a versatile and reliable tool for our analysis. We annotated images of Japanese macaques extracted from videos coming from Cédric Sueur and his team in various locations in Japan. The images selected for training the detector were collected at different Japanese study sites without specific differentiation. These images have a resolution of 6720x4480 pixels, captured using a Canon EOS 5D Mark IV camera. This high resolution enables us to observe intricate details of the animals, including the fine contours of individuals standing side by side. This superior visual quality is crucial for overcoming the challenges associated with annotating Japanese macaques in groups, given their thick fur. Our dataset used for training the detector initially consists of 1490 images, with annotations for 4149 individuals. Following augmentation using Roboflow®, our final dataset comprises 4290 images, with annotations for 11,945 individuals (examples in Figure 1). In this context, 'augmentation' refers to the process of creating new annotated images by duplicating and then modifying the images from the initial dataset (see Table 1 and Figure 2).

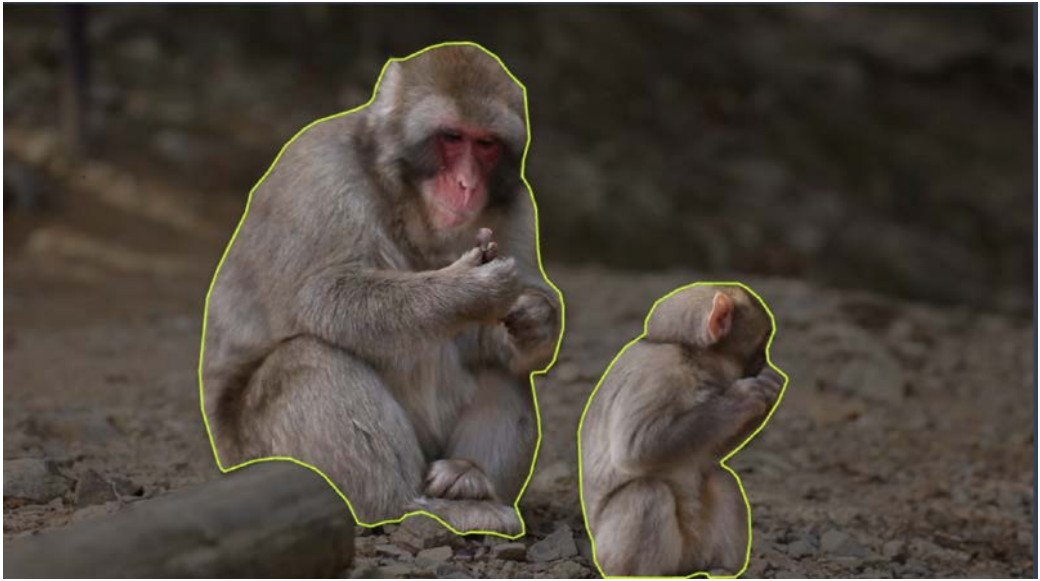

Figure 1: Picture featuring two Japanese macaques, with annotations highlighted in yellow using Roboflow®. Credit: Cédric Sueur

| Nature of image augmentation | Chosen parameter |
|---|---|
| 90° rotation | Clockwise |
| | Counterclockwise |
| | Reversed 180° |
| Cropping | Maximum zoom of 50% |
| Rotation | -35° to +35° |
| Shades of grey | Applied to 10% of images |
| Gaussian blur | Maximum 6.5 pixels |
| Background noise (random pixel replacement) | Maximum 4% |

Table 1: Parameters used for increasing the dataset used for training the **detector**.

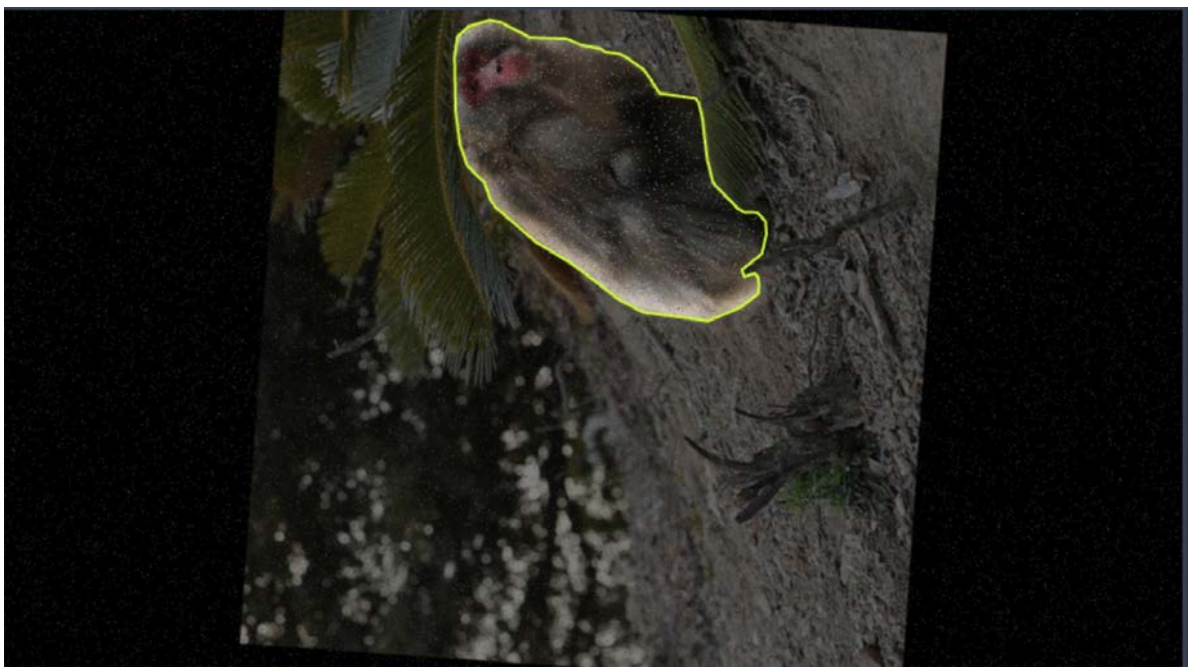

Figure 2: Example of augmentation applied to an image using Roboflow ®

*Training the Categoriser:* Once the Detector is proficient at identifying objects, the training of the Categoriser follows. This AI analyses the information provided by the detector to categorise and understand the actions or behaviours of the identified objects. Again, human supervisors play a crucial role in the training process by providing annotated examples of various actions or behaviours associated with the detected objects. The Categoriser learns from these examples, allowing it to categorise and interpret actions in new images. We focused on the analysis of two specific forms of stone play behaviours exhibited by Japanese macaques: 'Rub in hands' (RH) and 'Rub and roll on the ground' (RG). This choice is motivated by several factors. Firstly, as the first research team to apply LabGym1.0 to the study of primates, we opted for objectives that are simple and achievable within this timeframe. Secondly, the availability of video data depicting stone play behaviours is limited. This constraint reduces our ability to train the AI on less frequently observed forms of stone play. Therefore, we made the decision to focus our analysis on behaviours for which we have the most substantial amount of video data.

To train the Categoriser effectively, we utilised the 'behaviour example generator' integrated into LabGym1.0. This generator is a critical tool for preparing training data in an appropriate format. Here's how the process works:

—Selecting Videos: Initially, we selected a set of videos that depict the specific behaviours we are targeting for analysis.

—Determining Image Sequences: Next, we determined the minimum number of successive images required to adequately represent the desired behaviour. This step helps LabGym1.0 segment the chosen videos into sequences of images (animations) without sound. Each sequence effectively captures the behaviour we are interested in studying.

—Detecting Movements: These image sequences are then analysed by the detector mentioned earlier. This analysis generates a 'movement pattern' that highlights the movements of the macaque(s) present in the animation. In this pattern, the detected animal(s) at the beginning of the animation are represented in blue, and their colour gradually transitions to red as the animation progresses (see Figure 3). The movement pattern specifically represents the movements interpreted by the detector as those of a Japanese macaque in the image throughout the animation.

—Categoriser Training: The Categoriser is trained using all pairs of animations and their corresponding movement patterns. This comprehensive training process equips the Categoriser to conduct a more thorough and robust analysis of the targeted behaviours.

By following this procedure, we ensure that the Categoriser is well prepared to analyse and classify the behaviours of interest, in this case, 'Rub in hands' (RH) and 'Rub and roll on the ground' (RG), based on the movement patterns generated from the video data.

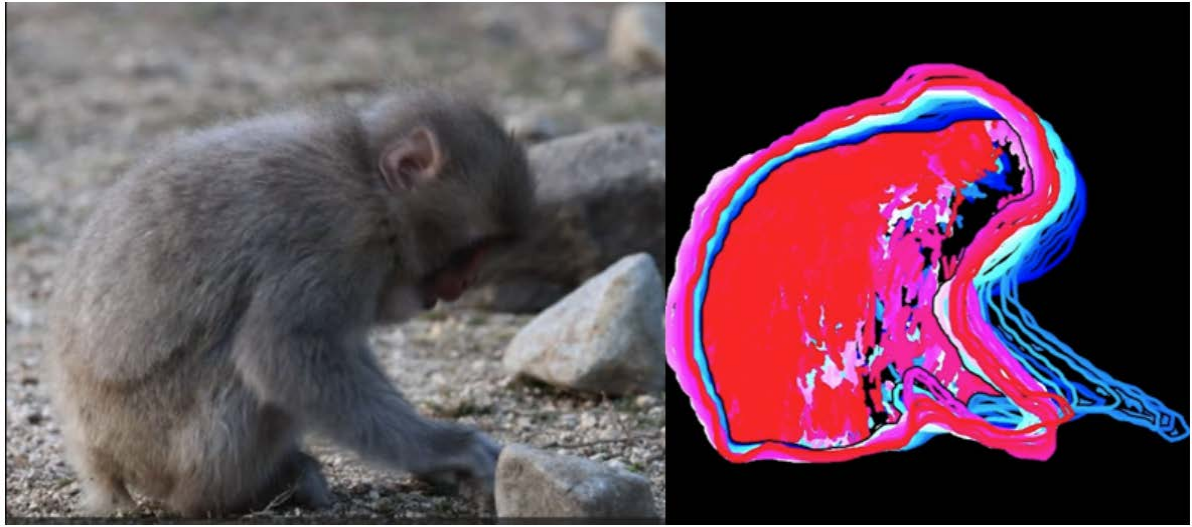

Figure 3: Illustration of an animation (left) and the associated movement pattern (right).

We have chosen to create sequences illustrating our targeted behaviours with a size of 45 images. While this number may seem somewhat arbitrary, it represents a compromise between sequences that are too short (e.g. 10 frames), which would not adequately capture the behaviour in a representative manner, and sequences that are excessively long (e.g. 100 frames or more), which could impact the quality of the Categoriser and increase the analysis time subsequently. Lengthy sequences require more processing time by our artificial intelligence, and they also carry the risk of including 'parasitic' behaviours unrelated to stone play. For instance, we aimed to avoid selecting sequences where a macaque engages in activities such as scratching itself immediately after playing with a stone. This precaution is taken to prevent any confusion in which the AI might associate unrelated gestures with the targeted behaviour.

To generate the training animations, we used videos featuring Japanese macaques engaged in stone play, primarily obtained from the monkey parks of Arashiyama and Shodoshima, where the practice of stone play has been observed for over 30 years. These videos have a more modest resolution, specifically 1920x1080 pixels, and are captured at a frame rate of 24 frames per second. Our dataset comprises 85 videos, with a combined duration of 54 minutes and 35 seconds. Following the processing of these videos, we have generated a total of 644 animations, each containing 45 frames, to illustrate the behaviour 'Rolling the stones on the ground.' Additionally, we have created 263 animations to illustrate the behaviour 'Rubbing the stone between the hands.'

*Estimation of the Detector Accuracy*

In LabGym1.0, there is not an integrated tool for assessing the accuracy of the Detector's Japanese macaque detection. This evaluation can only be conducted by a human analyst who meticulously reviews a substantial number of images processed by the detector. For the purpose of our evaluation, we considered a detection as accurate if it includes 95% or more of the surface area of the Japanese macaque within the silhouette drawn by LabGym1.0. If the surface area of the macaque detected falls below 95%, we referred to it as an 'under-detection'. On the other hand, if more than 5% of the silhouette drawn by LabGym1.0 included elements external to the macaque's body, we classified it as an 'over-detection'. This distinction helped us assess the precision of the detection, with under-detection indicating that the detector may have missed some parts of the macaque, and over-detection suggesting that the detector may have included non-macaque elements within the detection area (Figure 4, Supp mat 1).

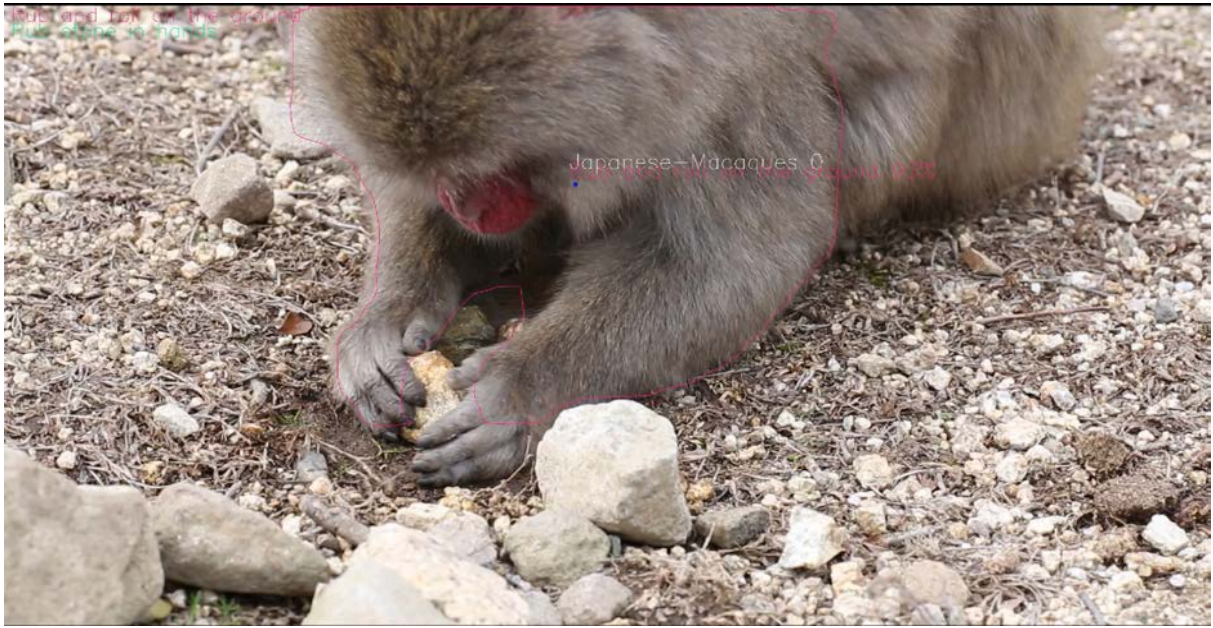

Figure 4: Image illustrating a case of underdetection of the Japanese Macaque. The posterior part of the macaque is not detected here.

Based on the advice provided by Yujia HU, the developer of LabGym1.0, we conducted a manual analysis of a total of 1,100 images. To do this, we extracted 100 images from each of the 11 videos that have been processed by the detector. This approach enabled us to observe variations in the precision of the detector across different parameters, including factors such as the macaque's position and its environment within the image. By manually analysing these images, we assessed the accuracy and consistency of the detector's performance across a diverse range of scenarios and contexts. This thorough evaluation process helped us gain valuable insights into the detector's capabilities and identify any areas where improvements may be needed.

*Estimation of the Categoriser Accuracy*

Unlike the Detector, the Categoriser offers quantitative data that provides information about the beginning and end of each stone game episode detected within a video. We analysed one video per behaviour using our most extensively trained Categoriser. Consequently, we were unable to calculate the average precision based on multiple analyses. Instead, we conducted a qualitative analysis of this single video. While a quantitative assessment across multiple analyses may have provided more comprehensive insights, a qualitative analysis of this video allowed us to gain valuable qualitative observations and insights into the stone game behaviours of Japanese macaques. This approach provided us with a foundational understanding of their behaviour in this specific context.

All of our work was conducted on a computation server equipped with 60 processors and 8 gigabytes of RAM. These specific parameters were chosen after conducting tests for training the detector on a dataset consisting of 15 annotated images (see Figure 5). This initial task enabled us to establish a balance between processing time and the efficient utilisation of the processors by our artificial intelligence.

## Results

*Computational requirements*

The processing time for the same task does not increase linearly with the number of processors utilised by the AI. Instead, a plateau is observed once the threshold of 60 virtual processors is surpassed. Beyond this point, the processing time remains constant (figure 5).

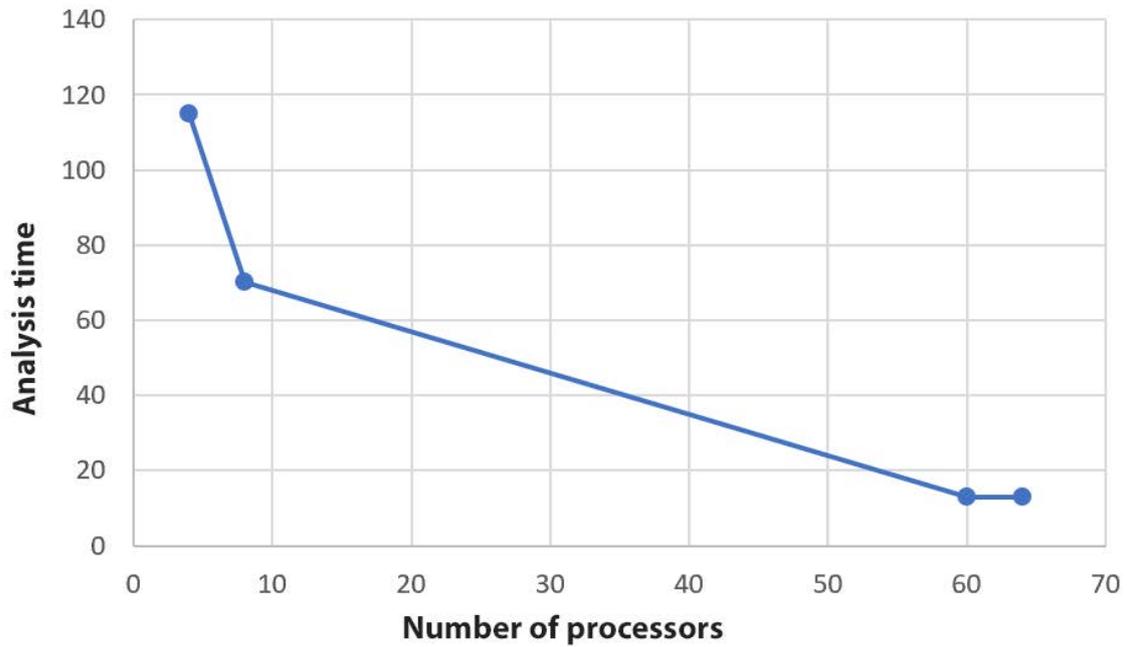

Figure 5: Processing time of a task depending on the number of virtual processors integrated into the calculation server.

*Detector accuracy*

Across 11 different videos (Figure 6), each featuring a single macaque in the image, the detector demonstrates an average accuracy of 77.3%, with a standard deviation of 18.88%. The accuracy of detection varies according to the characteristic of the video (table 2).

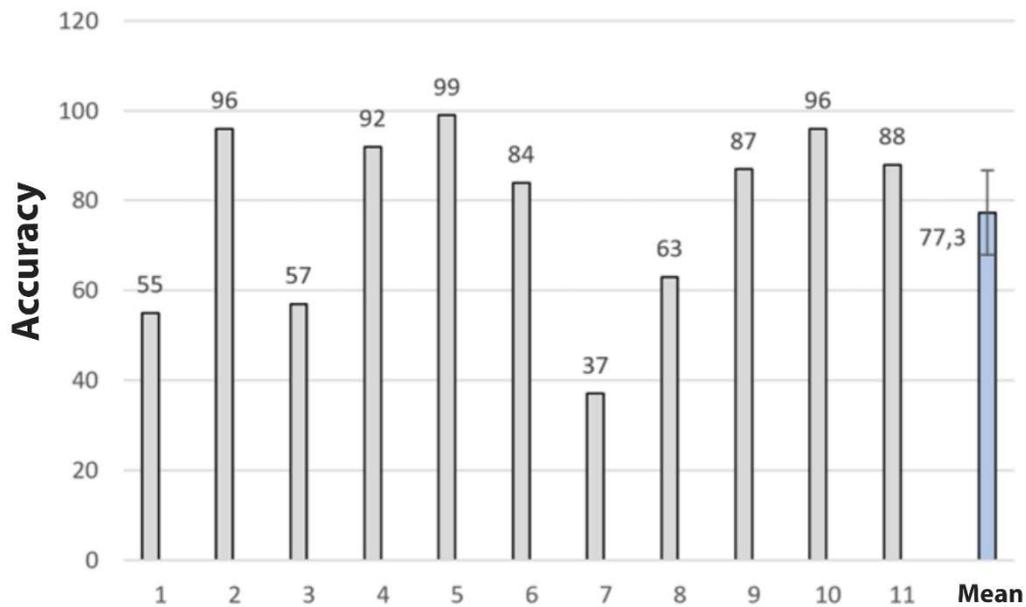

Figure 6: estimation of the accuracy of Japanese macaque detection by the detector. Each video analysed is indicated from 1 to 11.

| Video | Qualitative note on image capture and detection by *LabGym1.0* | Group average accuracy |
|---|---|---|
| 1 | In these cases, where the animal's body is partially outside the frame of the video and it is filmed from an overhead perspective, significant portions of the animal go undetected, resulting in a pronounced instance of under-detection. | 55% |
| 2,4,5,6,9,10,11 | In situations where the animal is entirely within the frame, and the recording is from a frontal or profile angle, with a significant colour contrast between the macaque and its surroundings, there may be slight under-detection observed, primarily affecting the tips of the animal's fingers. | 91.7% |
| 3,7,8 | When the animal is fully within the frame and the immediate surroundings, such as branches or stones, closely resemble the colour of the animal, there may be instances of over-detection where the environment in close proximity to the animal is also included in the macaque's detection. | 33% |

Table 2: Qualitative analysis of the content of the videos analysed by the **detector**.

*Categoriser accuracy*

We focused on a qualitative analysis of the results obtained for 'RG' (figure 7) and 'RH' (figure 8) behaviours within a 34-second video (Supp mat 2). This video was analysed using the most extensively trained Categoriser available, which yielded an estimated macaque detection accuracy of 83.7%.

The figure 7 illustrates the comparison between the probability that the ongoing action is 'Rub and roll on the ground' (RG), depicted in blue, and the actual action performed by the macaque (ground truth), annotated by the experimenter and shown in red. Notably, the AI recognises the behaviour of interest with a high confidence rates between 9.4 and 30.28 seconds, aligning with the anticipated results. However, it's important to highlight that the RG behaviour is not identified by the AI until 1.48 seconds into the sequence. The 'Rub in hands' (RH) behaviour occurs between 2.42 and 9.4 seconds in the video (figure 8), as annotated by the experimenter. However, the AI assigns a low probability to this behaviour, indicating poor detection by our Categoriser.

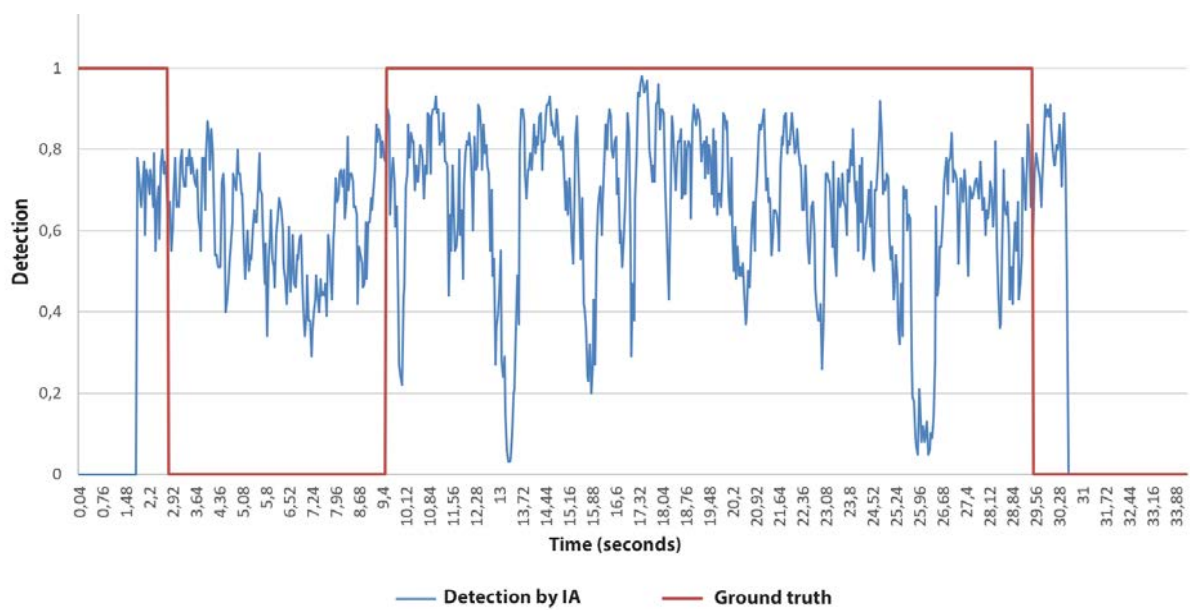

Figure 7: Categoriser results in detecting 'Rub and roll on the ground' behaviour

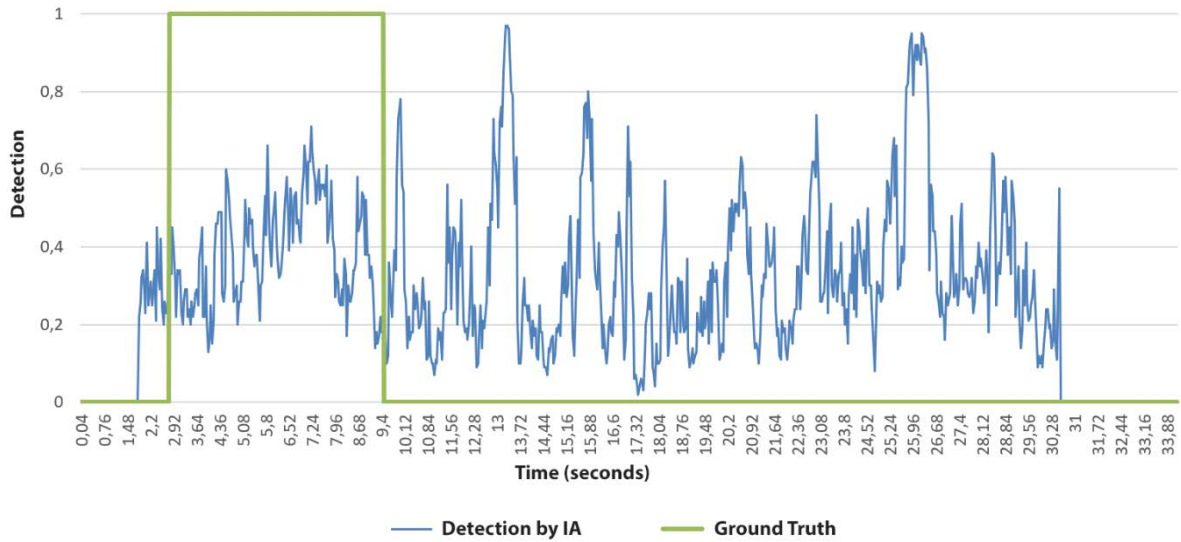

Figure 8: Categoriser results in detecting 'Rub in hands' behaviour

**Discussion**

As the first research team to employ LabGym1.0 for the analysis of primate behaviours, we encountered several challenges stemming from the scarcity of available data in the literature. Based on our results, we can identify factors that may contribute to the AI's lack of precision in tasks such as macaque detection and behaviour analysis.

Throughout this study, we observed certain limitations of LabGym1.0 in terms of task processing time. Notably, the AI's performance does not improve beyond the utilisation of 60 virtual processors, and it was not originally developed to leverage such a powerful computing server. Additionally, we noticed an uneven distribution of computational operations within our server. During intensive computing tasks, which could extend over several days, only 30 out of the 60 processors were consistently operating at full capacity, while the remaining 30 seldom exceeded 10%. Addressing this inconsistency in workload distribution may require collaboration with LabGym1.0 developers. Regarding macaque detection, the outcomes primarily depend on two critical factors: the macaque's size within the image and the colour of its immediate surroundings. When the entire macaque's body is not visible in the image, the detector's precision notably decreases, sometimes failing to detect up to half of the individual's body. Moreover, the majority of cases categorised as 'under-detection' correspond to the macaque's fingers being under-detected, likely due to their darker colour compared to the fur, which appears to confuse the detector. Additionally, the AI might interpret the environment close to the macaque as part of the macaque itself when the macaque is near branches. These observations highlight the need for further refinement and optimisation in AI performance, particularly in addressing the challenges posed by variations in image composition and environmental factors that affect detection accuracy. Reporting these issues to the LabGym1.0 developers could facilitate improvements in the system.

Nevertheless, under optimal conditions, the Detector exhibits an impressive average accuracy of 91.7%. This allowed us to train the Categoriser with the highest quality detection. During the Categoriser's analysis, we observed that the behavioural analysis for the 'Rub and roll on the ground' behaviour generally yielded better results than the 'Rub in hand' behaviour. In contrast, the 'Rub in hand' behaviour was consistently misdetected by the AI, even after employing very large artificial neural networks capable of deep analysis and trained with many videos. This subpar performance can be attributed to the scarcity of training data illustrating the 'Rub in hand' behaviour, as we had 644 training animations for the 'Rub and roll on the ground' behaviour compared to only 263 for the 'Rub in hand' behaviour. To advance this project, it is imperative to substantially increase the quantity of videos used for training the Categoriser. However, this means a considerable time on the field to score and collect the behaviours on videos.

While we could not evaluate the average precision of our most advanced Categoriser due to data limitations, its performance in analysing the 'Rub and roll on the ground' behaviour is deemed satisfactory. Indeed, the initial objective of this study was to develop AI capable of detecting and analysing Japanese macaque behaviours using LabGym1.0. We reached this aim and met our expectations regarding Japanese macaque detection, along with promising preliminary results for AI behavioural analysis. The insights gained from this project can serve as a foundational framework for our team and other researchers interested in pursuing similar work in this field.

Throughout this study, we encountered technical challenges stemming from the constraints of LabGym1.0, which was limited to analysing a single class of objects per video. However, it's worth noting that this constraint was lifted with the release of LabGym 2.0 on 07/12/2023 (Hu et al. 2023). Regrettably, we were unable to transition to LabGym 2.0 at that time due to resource constraints. The new version retains the same overall functionality as its predecessor but enables the simultaneous detection of multiple classes of objects and the analysis of their interactions. This development holds great promise for improving Japanese macaque detection, particularly by identifying the macaque's hands as an object, which will facilitate the analysis of stone play behaviours. In a broader context, LabGym2.0 opens up opportunities for the study of all behaviours involving tools in primates.

This first study serves as a vital stepping stone in the application of AI to ethology and behavioural science. It not only sheds light on the challenges and opportunities of using AI tools but also emphasises the importance of data collection, collaboration with developers, and staying updated with technological advancements. Ultimately, it contributes to the broader goal of advancing our understanding of primate behaviours and animal cognition through innovative research methodologies.

## Acknowledgments

We thank Yujia HU, the developer of LabGym1.0, for his help to apply LabGym to our study. Cédric Sueur was funded by an invited researcher fellowship from Kyoto University.## Supplementary material

*Supp mat 1: Video made with the first **Detector** (no behavioural analysis at this stage)*

https://youtu.be/CvSkN46lkOo7

*Supp mat 2: Video made with our most advanced **detector** and **Categoriser**.*

https://youtu.be/u58tjl4r8Vo